\documentclass[letterpaper]{article} %
\usepackage{aaai2027}  %
\usepackage[hyphens]{url}  %
\usepackage{graphicx} %
\usepackage{natbib}  %
\usepackage{caption} %
\usepackage{algorithm}
\usepackage{algorithmic}

\usepackage{newfloat}
\usepackage{listings}
\DeclareCaptionStyle{ruled}{labelfont=normalfont,labelsep=colon,strut=off} %
\floatstyle{ruled}
\newfloat{listing}{tb}{lst}{}
\floatname{listing}{Listing}

\usepackage{booktabs}
\usepackage{amsmath,amssymb}
\usepackage{booktabs}
\usepackage{multirow}
\usepackage{pifont}  %
\usepackage{hyperref}
\usepackage{cleveref}
\usepackage[table]{xcolor}
\usepackage{makecell}
\usepackage{marvosym}
\usepackage{graphicx}

\newcommand{\correspondingmark}{%
    \raisebox{-0.1ex}{\scalebox{1.05}{\Letter}}%
}

\title{UniNav: A Unified World-Action Diffusion Model for Visual Navigation}

\author{
    Changqing Zhou\textsuperscript{1}\quad
    Yueru Luo\textsuperscript{2}\quad
    Zeyu Jiang\textsuperscript{1}\quad
    Changhao Chen\textsuperscript{1\,\correspondingmark}
}

\affiliations{
    \textsuperscript{1}The Hong Kong University of Science and Technology (Guangzhou)\\
    \textsuperscript{2}The Chinese University of Hong Kong, Shenzhen\\[2pt]
    \texttt{czhou149@connect.hkust-gz.edu.cn}\quad
    \texttt{changhaochen@hkust-gz.edu.cn}
}

\newcommand{\modelname}{UniNav}

\begin{document}

\nocopyright

\maketitle

\begingroup
\renewcommand{\thefootnote}{\correspondingmark}
\footnotetext{Corresponding author.}
\endgroup

\begin{abstract}
Image-goal visual navigation is a fundamental capability for embodied agents. Existing navigation policies efficiently predict waypoint trajectories but lack visual foresight, while navigation world models can anticipate future observations but often require costly planning rollouts.
We present \textbf{\modelname}, a unified world-action model that generates future visual observations and continuous waypoint trajectories through a single diffusion process. Given history frames and a goal image, \modelname~jointly denoises visual and waypoint tokens within a single transformer, unifying future prediction and action generation in a shared framework.
To improve spatial grounding, we incorporate geometry-aware camera tokens. We also train on both trajectory-labeled navigation data and video-only data, enabling the model to benefit from diverse videos without waypoint annotations. Based on this unified framework, we introduce two variants: \textbf{\modelname{}-Full} jointly predicts interpretable future observations and their corresponding trajectories, while \textbf{\modelname{}-Fast} removes future-image tokens at inference for efficient trajectory prediction.
Experiments on navigation benchmarks show that \modelname~outperforms the strongest baseline in ATE across all datasets. With one-step inference, \modelname{}-Fast achieves a latency of $0.1$s without a substantial accuracy drop. Code will be released.
\end{abstract}

\section{Introduction}
\label{sec}

\begin{figure}[t]
\centering
\includegraphics[width=1.0\linewidth]{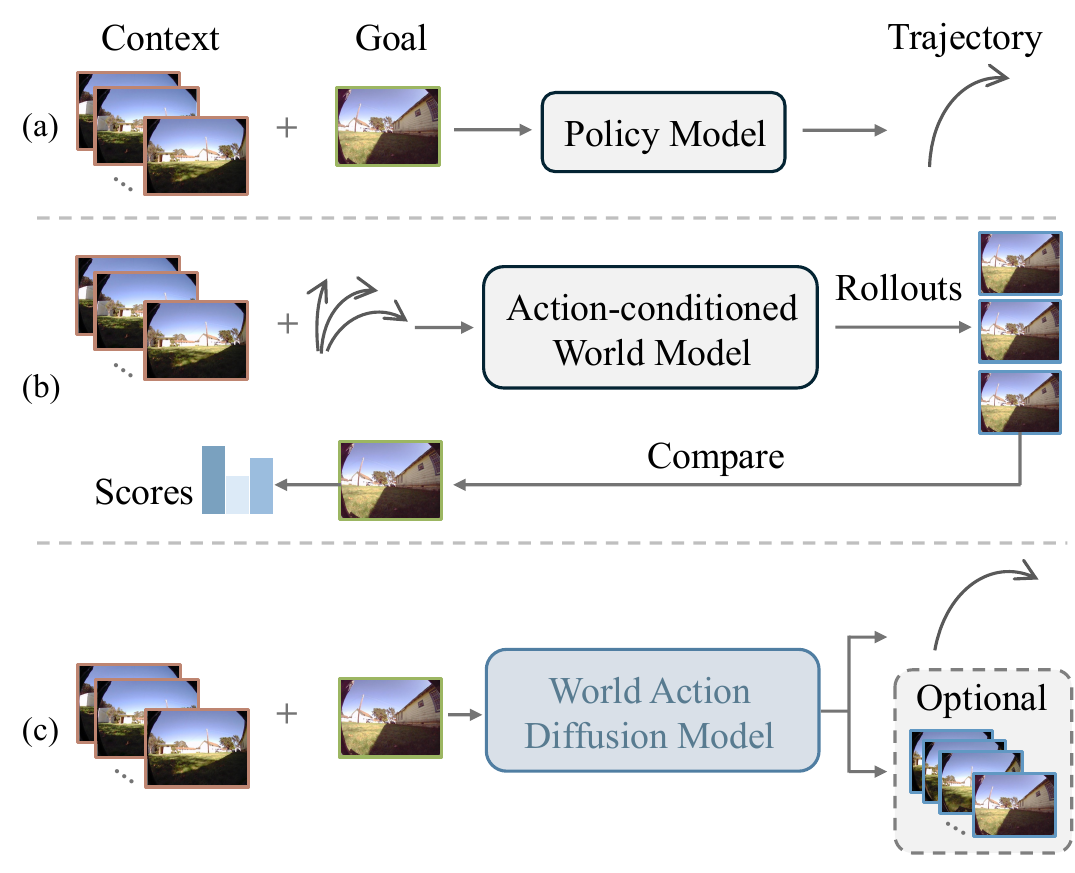}
\caption{
\textbf{Comparison of visual navigation paradigms.}
(a) Policy-only methods directly map the observation history and goal image to a waypoint trajectory, without explicitly modeling future observations.
(b) Action-conditioned navigation world models generate visual rollouts for candidate action sequences and select actions by comparing the imagined outcomes with the goal.
(c) Our unified world-action diffusion model jointly predicts executable waypoint trajectories and their corresponding egocentric visual futures within a single architecture, while supporting efficient action-only inference when visual prediction is not required.
}
\label{fig:teaser}
\end{figure}

Visual navigation is a fundamental capability for embodied agents, requiring robots to reach visually goals from egocentric observations while generalizing across environments and embodiments. Recent general-purpose navigation policy (action) models, such as GNM~\cite{gnm}, ViNT~\cite{vint}, and NoMaD~\cite{nomad}, learn transferable goal-reaching behaviors from large and diverse robot trajectory datasets. These models effectively predict local waypoint trajectories, establishing a promising foundation-model paradigm for robot visual navigation.
However, they formulate navigation primarily as direct observation-to-waypoint prediction. \textit{Human navigation, in contrast, is naturally accompanied by visual foresight: before moving toward a destination, we can anticipate how both the surrounding scene and the target will appear from future viewpoints.} Such anticipation provides an interpretable basis for action selection and enables reasoning about unseen parts of the environment. Waypoint-only policies do not explicitly model these future visual states, limiting both their interpretability and their ability to ground trajectory prediction in anticipated visual and geometric changes.
Navigation world models address this limitation by predicting future observations conditioned on candidate motion sequences~\cite{nwm,zhang2026rae,kim2026planning}. NWM~\cite{nwm}, for example, generates video rollouts for candidate trajectories and evaluates whether the imagined futures reach the goal. Subsequent works improve rollout quality using dense visual representations~\cite{zhang2026rae} or compact latent representations for efficient planning~\cite{kim2026planning}. Although these approaches provide explicit visual foresight, their generate-then-evaluate paradigm is computationally expensive because each candidate trajectory requires an independent rollout. This makes real-time closed-loop navigation increasingly difficult as the number of candidate trajectories grows.

World-action models offer a more efficient alternative by jointly modeling future world states and robot actions within a single generative process. Recent robotic manipulation systems demonstrate that unified video-action representations can effectively couple visual prediction with control generation~\cite{dreamzero,li2025unified,zhu2025unified,bi2026motus,li2026causal,pai2025mimic,liao2025genie,guo2026unified,li2026lightwamefficientworldaction}. Extending this paradigm to visual navigation, however, is challenging due to large-scale egocentric motion, long-horizon planning, and the need to jointly model future observations and continuous waypoint trajectories under geometric constraints.

Recent navigation world-action models have begun to address this challenge. UniWM~\cite{dong2025unified} interleaves action prediction and future-view generation in a memory-augmented autoregressive model, while WAM-Nav~\cite{yang2026wam} jointly generates trajectories and latent visual foresight with a diffusion transformer.
Nevertheless, existing methods either rely on costly visual rollouts, employ autoregressive architectures that scale poorly, or lack explicit metric geometric supervision. Furthermore, visual navigation is inherently multimodal, with multiple feasible trajectories leading to distinct future observations.
Diffusion models are well suited to capturing such joint world-action distributions. Combined with the strong priors of pretrained video diffusion models, they provide a scalable foundation for jointly predicting future observations and navigation actions. This motivates a geometry-aware unified world-action diffusion model that offers both interpretable visual foresight and efficient real-time waypoint prediction.

To this end, we propose \modelname, a unified world-action diffusion model for predictable image-goal visual navigation. Built upon the pretrained Wan2.1-1.3B video diffusion model~\cite{wang2025wan}, \modelname{} jointly predicts future egocentric observations and continuous waypoint trajectories. Given historical observations and a goal image, future visual latents, waypoint tokens, geometry-aware camera tokens~\cite{da3}, and register tokens~\cite{darcet2024vision,zhu2025unified,dinov2,wang2026vggt} are unified as token sequences and optimized under a shared flow-matching objective~\cite{lipman2022flow}, coupling visual foresight with executable navigation.
To improve spatial grounding, we introduce geometry-aware camera-token denoising, where camera tokens extracted by a frozen geometry foundation model~\cite{da3} provide 3D supervision for the shared world-action representation. To improve scalability, \modelname{} jointly leverages trajectory-labeled navigation data and video-only data, allowing abundant unlabeled videos to provide additional visual and geometric supervision without requiring waypoint annotations.
Finally, we develop two complementary variants. \modelname{}-Full jointly generates future observations and waypoint trajectories for interpretable visual foresight, while \modelname{}-Fast employs asymmetric attention~\cite{yuan2026fast,ye2026gigaworld,li2026metis} to decouple waypoint prediction from noisy image tokens, enabling efficient inference by discarding image tokens at test time.

Our contributions are summarized as follows:
\begin{itemize}
\item We propose a unified world-action diffusion framework that jointly models future egocentric observations and continuous waypoint trajectories using a pretrained video diffusion transformer.

\item We introduce geometry-aware camera-token denoising to transfer 3D structure from a frozen geometry foundation model into the shared world-action representation.

\item We incorporate video-only data, enabling additional visual and geometric supervision without requiring waypoint annotations for every sequence.

\item We develop two complementary variants: \modelname{}-Full for joint visual and trajectory generation, and \modelname{}-Fast for efficient waypoint prediction. Extensive experiments on multiple public visual navigation benchmarks demonstrate consistent performance improvements over existing methods.

\end{itemize}

\section{Related Work}
\label{sec:related}

\subsection{Image-Goal Conditioned Visual Navigation}

Learning-based visual navigation has evolved from dataset-specific policies toward general-purpose models trained on diverse multi-robot trajectories. GNM~\cite{gnm} established a shared goal-conditioned policy across robot platforms and environments, while ViNT~\cite{vint} introduced transformer-based temporal modeling for improved zero-shot transfer. NoMaD~\cite{nomad} further replaced deterministic action prediction with a diffusion policy, improving trajectory generation under multimodal navigation behaviors.
Subsequent work has focused on improving safety, efficiency, and geometric reasoning. NavDP~\cite{cai2025navdp} and NaviDiffusor~\cite{zeng2025navidiffusor} incorporate privileged information or cost guidance to improve obstacle avoidance. FlowNav~\cite{gode2025flownav} adopts conditional flow matching for faster inference and introduces monocular depth priors, while LoGoPlanner~\cite{peng2025logoplanner} integrates state estimation and metric 3D reasoning into a diffusion policy. Other approaches explicitly construct geometric representations, such as the 3D Gaussian Splatting maps used by BEINGS~\cite{meng2025beings}.
These methods primarily optimize action prediction and do not explicitly model the future visual consequences of the predicted trajectory. In contrast, \modelname{} retains an efficient waypoint-prediction path while jointly learning future visual dynamics through world-action training.

\subsection{Navigation World Models}

Navigation world models predict future egocentric observations conditioned on candidate actions, enabling planning through imagined rollouts. NWM~\cite{nwm} follows this paradigm by generating future video frames for action sequences and selecting actions through model predictive control. Although such visual rollouts provide interpretable predictions, evaluating multiple candidate trajectories requires repeated video generation and results in substantial computational cost~\cite{kim2026planning}.
Recent work improves the efficiency or consistency of this planning paradigm. MWM~\cite{yan2026mwm} separates structure learning from action-consistency post-training, while CompACT~\cite{kim2026planning} compresses observations into a small set of discrete latent tokens for faster planning. ReL-NWM~\cite{zhang2025efficient} and RAE-NWM~\cite{zhang2026rae} move world modeling from pixel space to DINOv2 feature space, reducing generation cost while preserving semantic and spatial structure. NavDreamer~\cite{huang2026navdreamer} instead generates navigation videos and extracts 3D waypoints from the imagined observations.
Despite these advances, action-conditioned world models generally require separate rollouts for candidate trajectories. \modelname{} avoids this bottleneck by directly modeling future observations and waypoint trajectories within a shared generative process, with a fast inference mode that predicts actions without generating future images.

\subsection{Unified World-Action Models}

Unified world-action models jointly learn visual dynamics and robot actions within a shared generative architecture. UVA~\cite{li2025unified} jointly models latent video and actions with decoupled decoding, while UWM~\cite{zhu2025unified} independently noises video and action tokens to support policy learning, dynamics modeling, and video generation. Fast-WAM~\cite{yuan2026fast} and GigaWorld-Policy~\cite{ye2026gigaworld} further show that visual prediction during training can improve action generation without requiring video synthesis at deployment.
This paradigm has been explored through diverse backbone and representation designs. Motus~\cite{bi2026motus} combines language, video, and action experts in a mixture-of-transformers architecture, while DreamZero~\cite{dreamzero}, Cosmos Policy~\cite{kim2026cosmos}, and LingBot-VA~\cite{lingbotva} adapt pretrained video-generation models for robot control. AIM~\cite{fan2026aim} and Multi-View VDP~\cite{li2026multi} further introduce structured representations based on spatial value maps and multi-view 3D-aware video latents, respectively.
Recent navigation methods have begun to adopt similar formulations. UniWM~\cite{dong2025unified} interleaves discrete action prediction and future-view generation in an autoregressive multimodal model with hierarchical memory, whereas WAM-Nav~\cite{yang2026wam} jointly models long-horizon actions and short-horizon latent visual foresight using a diffusion transformer.
In contrast, \modelname{} adapts a pretrained video diffusion transformer to image-goal navigation and jointly performs flow matching over continuous future-video latents, waypoint tokens, and camera-geometry tokens. This enables decoded RGB foresight, explicit geometric grounding, and efficient image-free waypoint inference within a single architecture.

\section{Method}
\label{sec:method}

\begin{figure*}[t]
\centering
\includegraphics[width=1.0\textwidth]{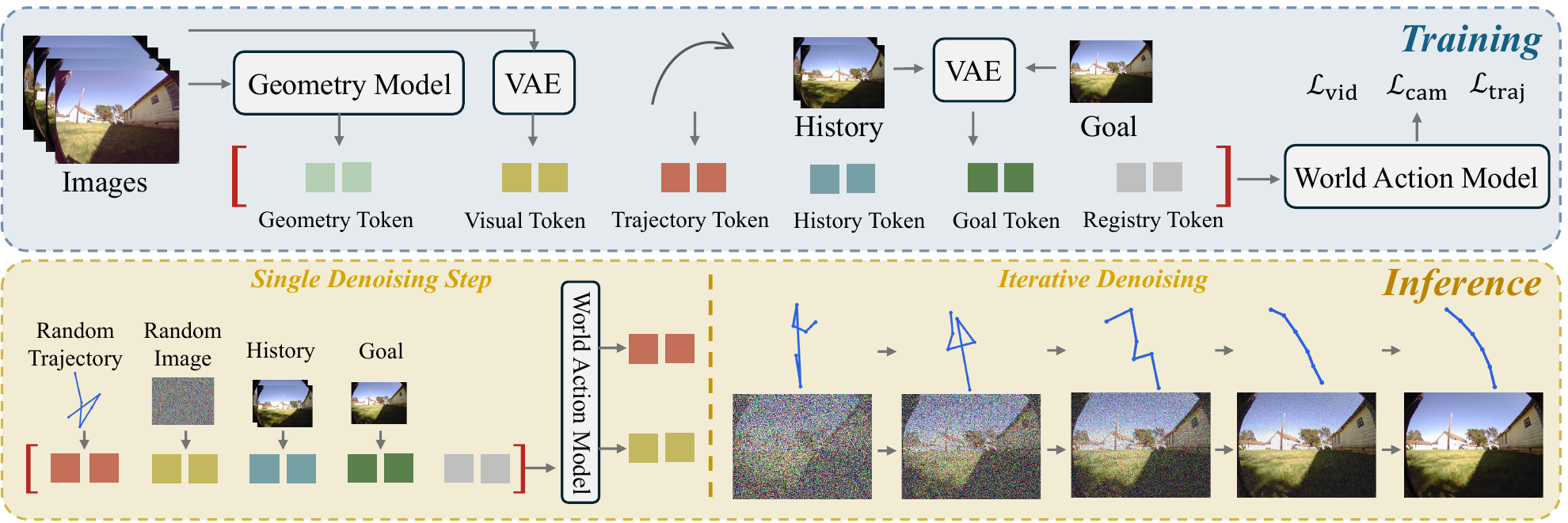}
\caption{
\textbf{Overview of \modelname{} training and inference}. During training, visual tokens from a pretrained VAE, camera-geometry tokens extracted by a frozen geometry model, trajectory tokens, history and goal tokens, and learnable register tokens are jointly processed by the world-action model. The model jointly denoises future-video, camera, and trajectory tokens using the corresponding objectives $\mathcal{L}_{\mathrm{vid}}$, $\mathcal{L}_{\mathrm{cam}}$, and $\mathcal{L}_{\mathrm{traj}}$. At inference, noisy future-image and trajectory tokens are initialized from random noise and iteratively denoised, conditioned on the history and goal observations, to jointly produce future visual observations and waypoint trajectories.
}
\label{fig:architecture}
\end{figure*}

\subsection{Overview}
\label{sec:overview}

We present \modelname, a geometry-aware world-action diffusion model for image-goal visual navigation. Given $C$ history frames ${I}_{t-C+1:t}=\{I_{t-C+1},\ldots,I_t\}$ and a goal image $I_g$, \modelname{} predicts a $K$-step local waypoint trajectory $\mathbf{a}_{t+1:t+K}\in\mathbb{R}^{K\times 3}$ and, optionally, the corresponding future egocentric observations $I_{t+1:t+K}$.
Each waypoint $\hat{\mathbf{a}}_{t+k}=(\Delta x_{t+k},\Delta y_{t+k},\Delta \psi_{t+k})$ specifies the relative translation and yaw change in the robot's local coordinate frame.

As illustrated in~\Cref{fig:architecture}, \modelname{} formulates navigation as a joint diffusion process over future visual observations, waypoint trajectories, and camera geometry.
During training, noisy visual, trajectory, and camera representations are jointly denoised under a unified flow-matching objective~\cite{lipman2022flow}, coupling visual foresight, geometric understanding, and executable action prediction within a single model.
At inference, the model starts from randomly initialized future-image and trajectories and iteratively refines them while conditioning on the history observations and goal image. This unified formulation enables \modelname{} to jointly generate future observations and waypoint trajectories, while also supporting efficient action-only inference by omitting future-image tokens.

\subsection{Geometry-Aware Token Representation}
\label{sec:world_action_tokenization}

To instantiate the joint diffusion process described above, \modelname{} represents the navigation context and prediction variables as a heterogeneous token sequence.
Specifically, the sequence consists of visual tokens, trajectory tokens, camera-geometry tokens, history and goal condition tokens, and learnable register tokens.

\paragraph{Visual tokens.}
The history frames, future frames, and goal image are encoded by a frozen video VAE~\cite{wang2025wan} and patchified into spatiotemporal latent tokens. The history tokens $\mathbf{h}_{\mathrm{hist}}$ and goal tokens $\mathbf{h}_{\mathrm{goal}}$ provide visual context, while the future-video tokens $\mathbf{h}_{\mathrm{pred}}$ are initialized from noised future latents and progressively denoised by the transformer.

\paragraph{Waypoint tokens.}
We represent the $K$-step local trajectory using $K$ continuous waypoint tokens, each corresponding to one waypoint in the robot's local coordinate frame. During training, each waypoint $\mathbf{a}_k$ is perturbed according to the flow-matching noise schedule, producing a noised trajectory state $\mathbf{z}_{\mathrm{traj},k}$. The noised waypoint is projected into the transformer hidden space using a lightweight MLP:
\begin{equation}
    \mathbf{h}_{\mathrm{traj},k}
    =
    \mathrm{MLP}_{\mathrm{traj}}(\mathbf{z}_{\mathrm{traj},k}).
\end{equation}
The resulting waypoint features are processed by the shared transformer and decoded into trajectory flow vectors~\cite{lipman2022flow} by a lightweight trajectory head.

\paragraph{Camera tokens.}
Image-goal navigation requires not only appearance matching but also an understanding of how the egocentric view changes under camera motion. Camera pose and 3D scene geometry therefore provide important cues for both future-view prediction and waypoint trajectory prediction~\cite{teed2021droid,wang2026vggt}. To incorporate such geometric information, we use a frozen geometry foundation model $G_{\xi}$~\cite{da3} to extract camera-aware target tokens for the $K$ future views:
\begin{equation}
    \mathbf{x}_{\mathrm{cam},t+1:t+K}
    =
    \mathrm{Norm}
    \left(
    G_{\xi}([\mathbf{o}_{t-C+1:t}, \mathbf{o}_{t+1:t+K}])_{t+1:t+K}
    \right),
\end{equation}
where
$\mathbf{x}_{\mathrm{cam},t+1:t+K}
=
\{\mathbf{x}_{\mathrm{cam},k}\}_{k=t+1}^{t+K}$
denotes the normalized camera tokens.
Each token provides a compact geometric target for its corresponding future view.
Following the unified flow-matching formulation, each clean camera token $\mathbf{x}_{\mathrm{cam},k}$ is perturbed with noise at a sampled timestep to obtain the noised camera state $\mathbf{z}_{\mathrm{cam},k}$.
The noised camera state is then projected into the transformer hidden space as
$\mathbf{h}_{\mathrm{cam},k}
    =
    \mathrm{MLP}_{c}(\mathbf{z}_{\mathrm{cam},k})$.
Denoising these camera tokens alongside the visual and waypoint representations encourages the model to capture metric scene structure.

\paragraph{Unified token sequence.}
As illustrated in~\Cref{fig:architecture}, the complete token sequence is organized as
\begin{equation}
[
\mathbf{h}_{\mathrm{cam}}
\mid
\mathbf{h}_{\mathrm{pred}}
\mid
\mathbf{h}_{\mathrm{traj}}
\mid
\mathbf{h}_{\mathrm{hist}}
\mid
\mathbf{h}_{\mathrm{goal}}
\mid
\mathbf{h}_{\mathrm{reg}}
].
\end{equation}
All token types are processed by the same denoising transformer, enabling geometric reasoning, future visual prediction, and waypoint prediction to interact within a shared representation space.
We further introduce a small set of learnable register tokens $\mathbf{h}_{\mathrm{reg}}$~\cite{darcet2024vision,li2025unified,dinov2} to facilitate global information exchange across heterogeneous token types. Visual tokens preserve the positional encoding of the pretrained video model~\cite{rope}, as they correspond to locations on the spatiotemporal latent grid. In contrast, trajectory, camera, and register tokens have no direct correspondence to image-grid locations and are therefore assigned identity positional encodings. This design avoids imposing artificial spatial coordinates on non-visual tokens while maintaining compatibility with the pretrained transformer.

\subsection{Unified Flow-Matching Objective}
\label{sec:flow_matching_objective}

We train \modelname{} with a unified flow-matching objective~\cite{lipman2022flow} over three target modalities: future-video latents, waypoint trajectories, and camera tokens. For each modality $m\in\{\mathrm{vid},\mathrm{traj},\mathrm{cam}\}$, let $\mathbf{x}_m$ denote the clean target, $\boldsymbol{\epsilon}_m$ Gaussian noise, and $\sigma_m$ a sampled noise level. The noised state and target velocity are defined as
\begin{equation}
    \mathbf{z}_m
    =
    (1-\sigma_m)\mathbf{x}_m
    +
    \sigma_m\boldsymbol{\epsilon}_m,
    \qquad
    \mathbf{u}_m
    =
    \boldsymbol{\epsilon}_m-\mathbf{x}_m.
\end{equation}
The shared transformer processes all modalities and predicts their velocities through lightweight modality-specific heads.

\paragraph{Video loss.}
Let $\mathbf{x}_{\mathrm{vid}}$ denote the clean future-video latents encoded by the frozen video VAE. We optimize
\begin{equation}
\mathcal{L}_{\mathrm{vid}}
=
\left\|
v_{\theta,\mathrm{vid}}(\mathbf{z},\sigma)
-
\mathbf{u}_{\mathrm{vid}}
\right\|_2^2 .
\end{equation}

\paragraph{Trajectory loss.}
For waypoint prediction, the clean target is the local trajectory
$\mathbf{x}_{\mathrm{traj}}=\mathbf{a}_{t+1:t+K}$.
We supervise the predicted trajectory velocity using a SmoothL1 loss:
\begin{equation}
\mathcal{L}_{\mathrm{traj}}
=
\sum_{k=1}^{K}
\sum_{d=1}^{3}
\mathrm{SmoothL1}
\left(
v_{\theta,\mathrm{traj},k,d}(\mathbf{z},\sigma),
\mathbf{u}_{\mathrm{traj},k,d}
\right),
\end{equation}
where $\mathbf{u}_{\mathrm{traj}}$ denotes the target flow velocity of the trajectory tokens.

\paragraph{Camera loss.} For geometric supervision, $\mathbf{x}_{\mathrm{cam}}=\mathbf{x}_{\mathrm{cam},t+1:t+K}$ denotes the normalized camera tokens extracted by the frozen geometry foundation model $G_{\xi}$. The camera-token loss is
\begin{equation}
\mathcal{L}_{\mathrm{cam}}
=
\left\|
v_{\theta,\mathrm{cam}}(\mathbf{z},\sigma)
-
\mathbf{u}_{\mathrm{cam}}
\right\|_2^2 .
\end{equation}
This auxiliary objective encourages the shared representation to preserve camera-aware geometric structure while keeping $G_{\xi}$ frozen.

\paragraph{Total objective.}
The overall training objective is
\begin{equation}
\mathcal{L}
=
\lambda_{\mathrm{vid}}m_{\mathrm{vid}}\mathcal{L}_{\mathrm{vid}}
+
\lambda_{\mathrm{traj}}m_{\mathrm{traj}}\mathcal{L}_{\mathrm{traj}}
+
\lambda_{\mathrm{cam}}m_{\mathrm{cam}}\mathcal{L}_{\mathrm{cam}},
\end{equation}
where $m_{\mathrm{vid}}$, $m_{\mathrm{traj}}$, and $m_{\mathrm{cam}}$ indicate whether the corresponding supervision is available for each sample.
Our training data consist of both trajectory-labeled navigation sequences and video-only sequences without action annotations. For navigation data, all three objectives are enabled, allowing the model to jointly learn visual dynamics, camera geometry, and executable waypoint prediction. For video-only data, we set $m_{\mathrm{traj}}=0$ and optimize only the video and camera-token objectives.
This mixed training strategy enables \modelname{} to benefit from large-scale and diverse video data without requiring additional waypoint annotations or introducing pseudo-trajectories.

\subsection{\modelname{}-Full and \modelname{}-Fast}
\label{sec:full_fast}

We instantiate \modelname{} as two model variants with different training and inference designs.

\paragraph{\modelname{}-Full.}
\modelname{}-Full uses the complete world-action token sequence defined in~\Cref{sec:world_action_tokenization} to jointly denoise future-video latents, waypoint trajectories, and camera tokens.
As illustrated in~\Cref{fig:architecture}, at each denoising step, the transformer predicts updates for the current noisy tokens conditioned on the history and goal observations. We iteratively apply this denoising process over multiple steps, ultimately obtaining an executable $K$-step trajectory together with its corresponding visual future.

\paragraph{\modelname{}-Fast.}
\modelname{}-Fast is trained as a separate, trajectory-efficient variant. It uses the complete token sequence during training but applies an asymmetric attention mask~\cite{yuan2026fast,ye2026gigaworld,li2026metis} that prevents all non-future tokens from attending to the noisy future-video tokens. The future-video tokens can still attend to the remaining tokens, retaining visual-foresight supervision without making trajectory denoising dependent on future-video representations.
At inference time, \modelname{}-Fast removes the future-video tokens and directly samples the waypoint trajectory, avoiding both future-video denoising and VAE decoding. This design reduces the sequence length and inference latency while retaining the same flow-matching formulation for trajectory prediction.

\section{Experiments}
\label{sec:experiments}

\begin{table*}[h]
\centering
\caption{Action prediction comparison on four navigation benchmarks. We report ATE and RPE, where lower values indicate better performance.}
\label{tab:main_action}
\small
\resizebox{0.85\textwidth}{!}{
\begin{tabular}{l cc cc cc cc cc}
\toprule
\multirow{2}{*}{Method}
  & \multicolumn{2}{c}{RECON}
  & \multicolumn{2}{c}{SACSoN}
  & \multicolumn{2}{c}{GO Stanford}
  & \multicolumn{2}{c}{SCAND}
  & \multicolumn{2}{c}{Avg.} \\
\cmidrule(lr){2-3}
\cmidrule(lr){4-5}
\cmidrule(lr){6-7}
\cmidrule(lr){8-9}
\cmidrule(lr){10-11}
  & ATE $\downarrow$ & RPE $\downarrow$
  & ATE $\downarrow$ & RPE $\downarrow$
  & ATE $\downarrow$ & RPE $\downarrow$
  & ATE $\downarrow$ & RPE $\downarrow$
  & ATE $\downarrow$ & RPE $\downarrow$ \\
\hline
\rowcolor{black!8}
\multicolumn{11}{c}{\textit{Action Only}} \\
ViNT~\cite{vint}
& 0.659 & 0.215
& 0.616 & 0.169
& 0.854 & 0.110
& 0.507 & 0.286
& 0.659 & 0.195 \\

NoMaD~\cite{nomad}
& 1.220 & 0.262
& 1.274 & 0.286
& 1.158 & 0.133
& 1.025 & 0.345
& 1.169 & 0.257 \\

NavDP~\cite{cai2025navdp}
& 0.952 & 0.259
& 0.738 & 0.439
& 0.840 & 0.242
& 0.649 & 0.413
& 0.795 & 0.338 \\

FlowNav~\cite{gode2025flownav}
& 1.021 & 0.351
& 1.091 & 0.435
& 1.058 & 0.141
& 0.950 & 0.527
& 1.030 & 0.364 \\

\hline
\rowcolor{black!8}
\multicolumn{11}{c}{\textit{World Model}} \\
NWM~\cite{nwm}
& 0.326 & \textbf{0.115}
& 0.653 & 0.325
& 0.895 & 0.120
& 0.497 & 0.258
& 0.593 & 0.205 \\

\textbf{\modelname{}-Fast (Ours)}
& 0.316 & 0.122
& 0.490 & 0.166
& \textbf{0.687} & \textbf{0.103}
& 0.474 & 0.250
& 0.492 & 0.160 \\

\textbf{\modelname{}-Full (Ours)}
& \textbf{0.314} & 0.121
& \textbf{0.464} & \textbf{0.156}
& 0.722 & 0.110
& \textbf{0.463} & \textbf{0.231}
& \textbf{0.491} & \textbf{0.155} \\
\bottomrule
\end{tabular}
}
\end{table*}

\begin{figure*}[h]
\centering
\includegraphics[width=\textwidth]{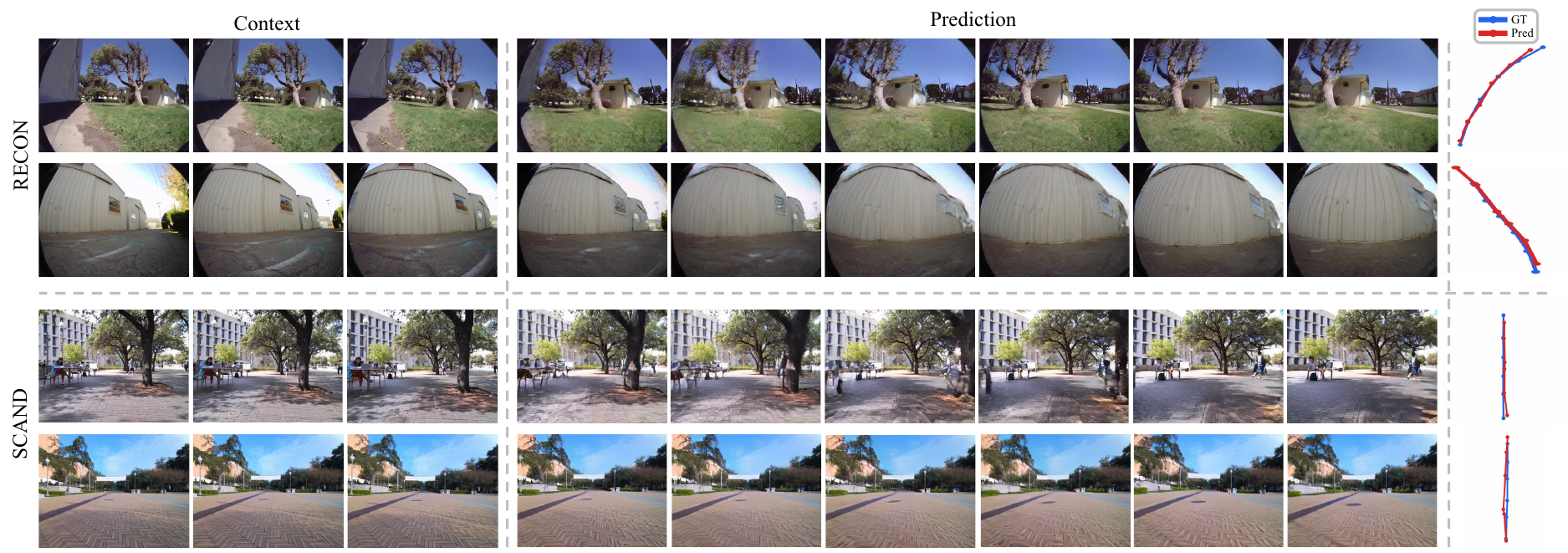}
\caption{
Qualitative joint prediction of future observations and waypoint trajectories.
The predicted trajectories are shown alongside the corresponding ground truth.
}
\label{fig:qualitative}
\end{figure*}

\subsection{Experimental Setup}

\paragraph{Datasets.}
We evaluate \modelname~on four real-world image-goal navigation datasets: RECON~\cite{recon}, SaCSoN~\cite{sacson}, GO Stanford~\cite{gostanford}, and SCAND~\cite{scand}. Each navigation sample consists of a short history of egocentric observations, a goal image, future observations, and local waypoint annotations. For mixed-data training, we additionally use video-only 3D datasets, including ScanNet~\cite{scannet}, DL3DV~\cite{dl3dv}, CityWalker~\cite{citywalker}, and LAVN~\cite{lavn}. These sequences supervise future-video and camera-token denoising but do not contribute to the action loss. Unless otherwise specified, we use $C=3$ history frames, predict $K=6$ future frames and waypoints, and resize all frames to $256\times192$.

\paragraph{Metrics.}
For waypoint prediction, we report Absolute Trajectory Error (ATE) and Relative Pose Error (RPE)~\cite{sturm2012evaluating}, where lower values indicate better performance. Future-frame quality is evaluated using PSNR, SSIM~\cite{ssim}, and LPIPS~\cite{lpips}. For inference efficiency, we report model latency and the number of function evaluations (NFE) under the same hardware setting as shown in~\Cref{tab:ablation_denoise_steps}.

\paragraph{Implementation details.}
We initialize \modelname~from the pretrained Wan2.1-T2V-1.3B video diffusion transformer~\cite{wang2025wan}. The video VAE remains frozen, while the diffusion transformer and newly introduced modules are fine-tuned. We use AdamW~\cite{adamw} with learning rates of $5\times10^{-6}$ for the pretrained transformer and $5\times10^{-5}$ for the newly introduced modules. Training is performed on four NVIDIA RTX 4090 GPUs for two epochs, with a batch size of 1 per GPU and gradient accumulation over 4 iterations. The loss weights are set to $\lambda_{\mathrm{vid}}=1$, $\lambda_{\mathrm{act}}=5$, and $\lambda_{\mathrm{cam}}=0.01$. Unless otherwise specified, inference uses 2 sampling steps.

\subsection{Main Results}

We compare \modelname~with two groups of methods: action-only policies, including ViNT~\cite{vint}, NoMaD~\cite{nomad}, NavDP~\cite{cai2025navdp}, and FlowNav~\cite{gode2025flownav}; and the navigation world-model baseline NWM~\cite{nwm}. We evaluate two variants of our model. \textbf{\modelname{}-Full} keeps future-image tokens and jointly predicts future frames and waypoint actions, while \textbf{\modelname{}-Fast} removes future-image tokens and skips future-frame decoding during inference.

To ensure a fair comparison with prior navigation methods, all results in the main comparison are obtained using \textbf{only trajectory-labeled navigation data}, without additional video-only training data.
\Cref{tab:main_action} reports waypoint prediction results on four real-world navigation benchmarks. All results of \modelname~use the default two-step sampler. \modelname~achieves the best ATE on all datasets, with reductions of $3.7\%$ on RECON, $24.7\%$ on SaCSoN, $18.2\%$ on GO Stanford, and $6.8\%$ on SCAND compared with the strongest baseline on each benchmark. \modelname~also delivers competitive RPE across datasets, showing that the improvement in trajectory accuracy is consistent across different evaluation metrics. These results indicate that jointly modeling future visual dynamics and waypoint actions improves waypoint prediction across diverse real-world navigation settings. Moreover, \modelname{}-Fast maintains strong accuracy while removing future-frame decoding, providing a more efficient inference mode for visual navigation.

\subsection{Ablation Studies}

\begin{figure*}[h]
\centering
\includegraphics[width=\textwidth]{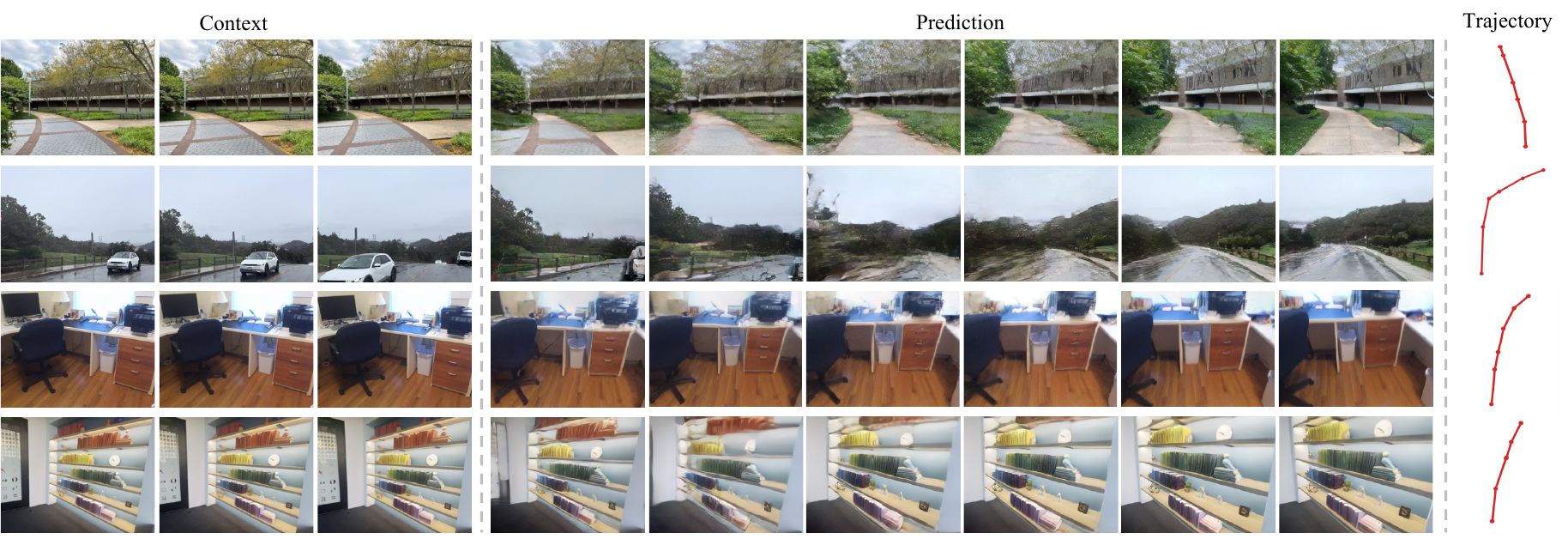}
\caption{
Qualitative future-frame predictions on video-only datasets trained without waypoint annotations.
}
\label{fig:qua_video_only}
\end{figure*}

We conduct ablation studies on RECON~\cite{recon} to isolate the key design choices of \modelname{}.
In the main paper, we investigate the effects of camera-token geometry supervision and mixed training, as well as the accuracy-efficiency trade-offs of different model configurations.
Additional analyses are provided in the Appendix, including visual prediction performance, alternative world-action prediction designs, and zero-shot evaluation on real world robot-collected data.

\begin{table}[h]
\centering
\caption{
Effect of camera-token geometry supervision on RECON.
Clean camera-token conditioning is an oracle setting that extracts camera tokens from ground-truth future frames and is therefore unavailable in practical inference.
}
\label{tab:ablation_cam_constraint}
\small
\begin{tabular}{lcc}
\toprule
Geometry design & ATE $\downarrow$ & RPE $\downarrow$ \\
\hline
\noalign{\smallskip}
Without camera tokens & 0.321 & 0.125 \\
Clean camera-token conditioning & 0.299 & 0.102 \\
Camera-token denoising (ours) & 0.314 & 0.121 \\
\bottomrule
\end{tabular}
\end{table}

\paragraph{Effect of Camera-Token Geometry Supervision.}
\Cref{tab:ablation_cam_constraint} compares three ways of incorporating camera geometry: removing camera tokens, conditioning on clean camera tokens extracted by DA3~\cite{da3}, and treating camera tokens as an auxiliary denoising target.
Clean camera-token conditioning achieves the lowest ATE and RPE, confirming that future-view geometry provides useful guidance for trajectory prediction. However, this result represents an oracle setting: the camera tokens are extracted from ground-truth future frames, which are unavailable during practical inference.
In contrast, our camera-token denoising design does not require access to future observations at test time and still improves over the model without camera tokens. These results demonstrate that explicitly learning to denoise camera tokens provides effective geometric supervision while preserving a fully deployable inference pipeline.

\begin{table}[h]
\centering
\caption{
Effect of video-only data and camera-token supervision on RECON.
}
\label{tab:ablation_mixed_data}
\small
\setlength{\tabcolsep}{4pt}
\begin{tabular}{llcc}
\toprule
Cam. denoising & Data & ATE $\downarrow$ & RPE $\downarrow$ \\
\hline
\noalign{\smallskip}
\multirow{3}{*}{No}
& Navigation & 0.321 & 0.125 \\
& + Video-only & 0.330 & 0.125 \\
& & +0.009 & 0.000 \\
\hline
\noalign{\smallskip}
\multirow{3}{*}{Yes}
& Navigation & 0.314 & 0.121 \\
& + Video-only & 0.302 & 0.109 \\
& & -0.012 & -0.012 \\
\bottomrule
\end{tabular}
\end{table}

\paragraph{Effect of Mixed Navigation and Video-Only Training.}
\Cref{tab:ablation_mixed_data} examines how the proposed camera-token geometry supervision affects the use of video-only training data.
Trajectory-labeled navigation samples supervise video, trajectory, and camera-token denoising, whereas video-only samples provide only video and camera-token supervision.
Without camera-token denoising, adding video-only data increases ATE by $0.009$, suggesting that visual prediction supervision alone does not effectively transfer to waypoint prediction.
In contrast, when camera-token denoising is enabled, adding video-only data reduces both ATE and RPE by $0.012$, yielding the best overall performance of $0.302/0.109$.
These results demonstrate that our camera-token geometry supervision provides an effective geometric connection between visual dynamics and navigation actions, allowing \modelname{} to benefit from video-only data.

\begin{table}[h]
\centering
\caption{
Accuracy and efficiency profiles under different denoising steps on RECON.
Latency is measured on a single NVIDIA RTX 4090 GPU.
}
\label{tab:ablation_denoise_steps}
\small
\scalebox{0.9}{
\begin{tabular}{lccccc}
\toprule
Model & Steps & ATE $\downarrow$ & RPE $\downarrow$ & Latency (s) $\downarrow$ & NFE \\
\hline
\noalign{\smallskip}
ViNT & -- & 0.659 & 0.215 & 0.015 & 1 \\
NoMaD & 10 & 1.220 & 0.262 & 0.051 & 10 \\
FlowNav & 10 & 1.021 & 0.351 & 0.058 & 10 \\
\hline
\noalign{\smallskip}
NWM & 50 & 0.326 & 0.115 & $\sim$900 & 12{,}000 \\
\hline
\noalign{\smallskip}
\multirow{6}{*}{\modelname{}-Full}
& 1 & 0.312 & 0.121 & 0.192 & 1 \\
& \textbf{2} & 0.314 & 0.121 & 0.389 & 2 \\
& 5 & 0.310 & 0.105 & 0.810 & 5 \\
& 10 & 0.308 & 0.099 & 1.815 & 10 \\
& 20 & 0.299 & 0.097 & 3.353 & 20 \\
& 50 & 0.312 & 0.103 & 6.112 & 50 \\
\hline
\noalign{\smallskip}
\multirow{6}{*}{\modelname{}-Fast}
& 1 & 0.318 & 0.123 & 0.109 & 1 \\
& \textbf{2} & 0.316 & 0.122 & 0.212 & 2 \\
& 5 & 0.312 & 0.120 & 0.535 & 5 \\
& 10 & 0.300 & 0.105 & 1.040 & 10 \\
& 20 & 0.313 & 0.103 & 2.110 & 20 \\
& 50 & 0.331 & 0.107 & 5.360 & 50 \\
\bottomrule
\end{tabular}
}
\end{table}

\paragraph{Model Profile.}
\Cref{tab:ablation_denoise_steps} reports the accuracy--efficiency profiles of \modelname{} under different numbers of denoising steps.
With the default two-step sampler, \modelname{}-Full and \modelname{}-Fast achieve similar trajectory accuracy, while \modelname{}-Fast reduces latency from $0.389$\,s to $0.212$\,s by removing future-image tokens and video decoding.
With one-step sampling, \modelname{}-Fast further reduces latency to $0.109$\,s while retaining strong accuracy, providing a practical operating point for online waypoint prediction.
Increasing the number of denoising steps can moderately improve accuracy, but substantially increases latency and does not yield monotonic gains.
Compared with the planning-based NWM~\cite{nwm}, both variants require far fewer function evaluations and complete inference within a much shorter time, making \modelname{} more suitable for efficient navigation deployment.

\subsection{Qualitative Results}

\Cref{fig:qualitative} shows coherent future observations and waypoint trajectories across diverse navigation environments.
\Cref{fig:qua_video_only} presents plausible future predictions on video-only datasets without waypoint annotations.
The Appendix provides additional visualizations for both settings, together with zero-shot results on our robot-collected real-world data.

\section{Conclusion}
\label{sec:conclusion}

We presented \modelname{}, a geometry-aware world-action diffusion model that unifies future observation prediction and waypoint generation for image-goal visual navigation.
By jointly denoising visual, trajectory, and camera-geometry tokens, \modelname{} couples visual foresight with executable action prediction in a single model.
Its Full variant provides interpretable visual predictions, while the Fast variant enables efficient action inference by removing future-image tokens.
Moreover, mixed training with trajectory-labeled navigation data and video-only data allows the model to benefit from heterogeneous visual experience without requiring action annotations for every sequence.

\noindent\textbf{Limitations and Future Work.}
Our current model focuses on short-horizon local waypoint prediction and does not explicitly address long-horizon planning.
In addition, our evaluation is primarily conducted in offline settings, and broader closed-loop deployment remains to be explored.
Future work will investigate integration with long-horizon planners and extensive closed-loop evaluation across diverse robots and real-world environments.

\bibliography{aaai2027}

\end{document}